%% file: main.tex
\documentclass[11pt]{article}
\usepackage[preprint]{acl}

\usepackage{times}
\usepackage{latexsym}
\usepackage[T1]{fontenc}
\usepackage[utf8]{inputenc}
\usepackage{microtype}
\usepackage{inconsolata}
\usepackage{graphicx}
\usepackage{booktabs}
\usepackage{multirow}
\usepackage{amsmath}

\title{LitArc-Bench: Disentangling Text-Specific Reader-Response\\
Dynamics from Position and Lexical Affect}

\input{authors}
\hypersetup{
  pdftitle={LitArc-Bench: Disentangling Text-Specific Reader-Response Dynamics from Position and Lexical Affect},
  pdfauthor={Mingming Su, Yushan Wei, Sen Song, Xinche Zhang}
}

\begin{document}
\maketitle

\begin{abstract}
\input{sections/abstract}
\end{abstract}

\section{Introduction}
\input{sections/introduction}

\section{Related Work}
\input{sections/related_work}

\section{LitArc-Bench}
\input{sections/dataset}

\section{Human Reliability and Reference Arcs}
\input{sections/human_reference}

\section{LLM Evaluation Protocol}
\input{sections/experiments}

\section{Results}
\input{sections/results}

\section{Analysis}
\input{sections/analysis}

\section{Conclusion}
\input{sections/conclusion}

\section*{Limitations}
\input{sections/limitations}

\section*{Ethical Considerations}
\input{sections/ethics}

\bibliography{references}

\appendix
\input{sections/appendix}

\end{document}

%% file: authors.tex
\author{
\textbf{Mingming Su}\textsuperscript{1,*}\quad
\textbf{Yushan Wei}\textsuperscript{2,*}\quad
\textbf{Sen Song}\textsuperscript{3}\quad
\textbf{Xinche Zhang}\textsuperscript{3,\textdagger}
\\[2pt]
{\normalfont\normalsize\textsuperscript{1}Capital University of Economics and Business}
\\
{\normalfont\normalsize\textsuperscript{2}Shanghai International Studies University}
\\
{\normalfont\normalsize\textsuperscript{3}Tsinghua University}
\\[2pt]
{\normalfont\small\textsuperscript{*}Equal contribution.\quad
\textsuperscript{\textdagger}Corresponding author.}
}

%% file: sections/abstract.tex
When model predictions track an unfolding human affect trajectory, alignment
may reflect narrative position, affective words, or text-specific
reader-response dynamics. We introduce LitArc-Bench to separate these
explanations: 39 Chinese-language literary excerpts segmented into seven
checkpoints, with 430 questionnaires from 81 adults in two classroom groups
(9,020 VAD ratings). We compare three future-blind model runs with
leave-one-text-out position priors, lexical affect, and human--human
references. Position alone reaches $r=.470$ for arousal, making raw alignment
insufficient. Yet all six V/A transition advantages over a joint
position--lexicon control have text-bootstrap intervals above zero.
Cross-fitted models match the correct human text better than 38 mismatched
texts in all nine model--dimension cells ($p<.0001$); a text-invariant negative
control returns chance ($p=1$). On human transitions at least 0.5 points
against the position trend, models recover 69--76\% of valence directions and
53--68\% for arousal and dominance. Under crossed participant--text
resampling, all nine identity intervals remain above zero; joint-control
advantages are robust for valence and directional but less certain for
arousal. High shape agreement still coexists with scale bias. LitArc-Bench
identifies which components of literary reader response are recoverable beyond
shallow explanations and why shape, calibration, and human variability require
separate evaluation.

%% file: sections/introduction.tex
Readers do not experience literary emotion as isolated labels. A passage may
begin in calm description, move into tension or loss, and end with relief,
estrangement, or unresolved ambiguity. This temporal shape is central to
literary reading, but largely absent from standard sentiment benchmarks, which
typically assign a single polarity or emotion category to a short text.

High model--human correlation in this setting is ambiguous. A system can look
successful without following the target passage: literary excerpts may share a
generic buildup over segment positions, and affective words can approximate
local valence or arousal. Conversely, a passage can reverse those cues---a
threat is resolved despite negative words, or a formally positive announcement
creates coercive unease. The computational question is therefore not simply
whether an LLM can emit plausible ratings, but which part of its alignment
remains when position and lexical affect are made explicit competitors.

We represent every excerpt as a seven-point valence, arousal, and dominance
arc. Three future-blind LLM conditions see only the textual prefix available
at each rating point. A leave-one-text-out position prior sees no target text,
and a Chinese affect lexicon sees local words without modeling the unfolding
episode. All predictors are evaluated against the same densely measured human
arc. We distinguish global shape, adjacent changes, and numeric calibration,
then place model agreement against ICC, split-half, and leave-one-reader-out
human references. This design turns model outputs into diagnostic probes of
recoverable reader-response structure rather than a provider leaderboard.

We make three contributions. First, we introduce LitArc-Bench, comprising 39
Chinese-language literary excerpts and 430 reader--text questionnaires from 81
adult readers. Participants rated one to ten excerpts each (median six), while
each text received a multi-reader panel, yielding 9,020 observed ratings and
reconstructable human--human references. Second, we introduce a
reliability-aware decomposition of apparent alignment into position, lexical
affect, text-specific shape and change, and scale calibration. Third, we show
that generic position explains substantial arousal alignment, yet model
change advantages survive a joint position--lexicon control and the correct
human text remains identifiable after cross-fitted removal of shallow effects
from the human target. This identity result replicates across three models,
survives crossed participant--text uncertainty, and disappears for a
text-invariant negative control. Models also recover a majority of strong counter-position
changes, while global dominance shape and numeric calibration remain less
stable. The contribution is thus a measurement account of what literary
model--human alignment captures, not a claim that correlation alone
demonstrates human-like reading.

\begin{figure*}[t]
\centering
\includegraphics[width=.96\textwidth]{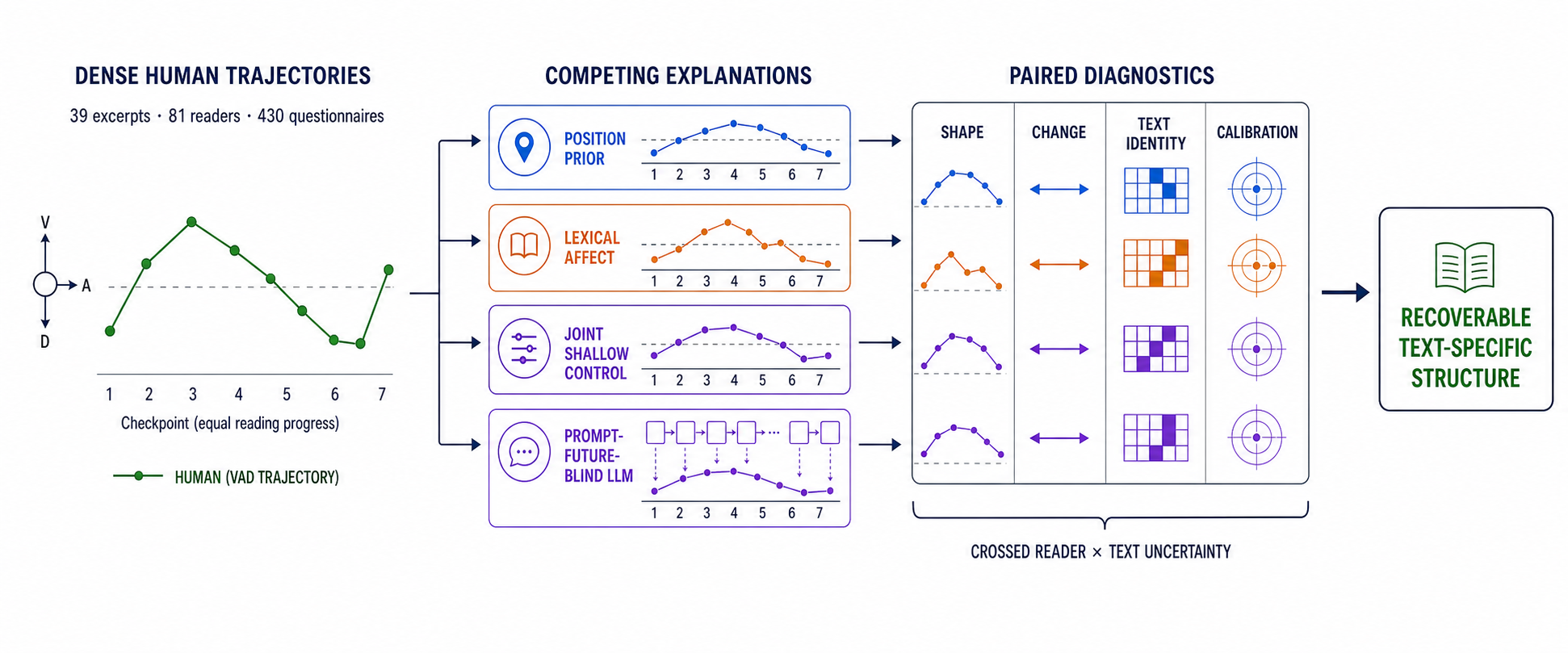}
\caption{Study design and diagnostic logic. Each human reference is a
seven-checkpoint VAD trajectory. Position, lexical affect, a joint
position--lexicon control for V/A, and prompt-future-blind LLMs are evaluated
against the same reference using trajectory shape, adjacent change,
correct-text identity, and numeric calibration. Uncertainty is propagated
across both readers and texts.}
\label{fig:benchmark-overview}
\end{figure*}

%% file: sections/related_work.tex
\subsection{Dimensional emotion and perspective}

Valence, arousal, and dominance (VAD) represent affect continuously rather
than as a single polarity or categorical emotion
\citep{russell1980circumplex,bradley-lang-1994-sam,mohammad2018obtaining}.  This representation has
supported resources ranging from word norms to sentence-level corpora.
EmoBank, for example, annotates 10,000 English sentences and explicitly
distinguishes the emotion expressed by a writer from that evoked in a reader
\citep{buechel-hahn-2017-emobank}.  Related experiments show that annotation
quality depends on both perspective and domain
\citep{buechel-hahn-2017-readers}.  Chinese dimensional resources have also
modeled valence and arousal at sentence and context level, including the role
of irony \citep{huang-etal-2022-dimensional}.  LitArc-Bench shares this
dimensional and reader-centered view, but differs in treating successive
ratings from the same reading episode as a trajectory rather than independent
sentence labels.

\subsection{Emotion in literary narrative}

Computational work on literary emotion has mainly characterized affect
represented \emph{in} a text.  REMAN annotates emotion triggers together with
experiencers, targets, and causes in fiction \citep{kim-klinger-2018-feels}.
Reader-response research instead treats fiction as a cognitive and emotional
simulation enacted by readers \citep{oatley1999fiction}, motivating direct
measurement of evoked rather than merely expressed affect.
At the book scale, lexical sentiment has been used to derive recurring emotional
arc shapes \citep{reagan2016emotional}; EmotionArcs extends this approach to
more than 9,000 English novels and relates automatically derived arcs to
reader appreciation \citep{ohman-etal-2024-emotionarcs}.
Character-specific work shows that narration and dialogue can follow distinct
emotion dynamics \citep{vishnubhotla-etal-2024-emotion}.  CR4-NarrEmote
collects open-vocabulary narrative-emotion judgments from
thousands of citizen readers and maps them to categorical and VAD
representations \citep{piper-budac-2025-cr4}.  These resources establish that
narrative emotion is both scalable and reader-relevant.  Our complementary
focus is controlled sequential response: 39 Chinese-language excerpts are
read in seven increments, every point is a contemporaneous human rating, and
the focal model is evaluated under the same information boundary.
For copyrighted sequential corpora, annotations can be distributed with
non-reversible hashes for local alignment by lawful source holders
\citep{amalvy-etal-2026-overcoming}. We use this pattern, releasing annotations,
exact hashes, and an alignment utility rather than passages.

\begin{table*}[t]
\centering
\footnotesize
\setlength{\tabcolsep}{3.2pt}
\begin{tabular}{p{0.17\textwidth}p{0.15\textwidth}p{0.21\textwidth}ccc}
\toprule
Resource & Language / unit & Primary affect signal &
\shortstack{Same-reader\\sequence?} &
\shortstack{Human arc\\reference?} &
\shortstack{Future-blind\\model test?} \\
\midrule
EmoBank \citep{buechel-hahn-2017-emobank}
& English / sentence & Reader/writer VAD & No & No & No \\
CDVAI \citep{huang-etal-2022-dimensional}
& Chinese / sentence+context & VA and irony & No & No & No \\
EmotionArcs \citep{ohman-etal-2024-emotionarcs}
& English / novel bins & Automated text emotion & No & No & No \\
CR4-NarrEmote \citep{piper-budac-2025-cr4}
& English / passage & Open-vocabulary reader report & No & No & No \\
\textbf{LitArc-Bench}
& \textbf{Chinese / 7 increments} & \textbf{Incremental reader VAD}
& \textbf{Yes} & \textbf{Yes} & \textbf{Yes} \\
\bottomrule
\end{tabular}
\caption{Closest affective and narrative resources differ from LitArc-Bench in
measurement design rather than overall quality or scale. ``Same-reader sequence''
requires successive ratings from the same reading episode, not merely labels
at several textual locations. Human--human arc reference denotes a trajectory
statistic reported on the same trajectory metric as model correlation, while
retaining a different noise structure.}
\label{tab:resource-comparison}
\end{table*}

\subsection{Human variation and simulated readers}

For subjective tasks, averaging annotations can obscure systematic differences
among annotators \citep{davani-etal-2022-dealing}.  Human label variation is
therefore part of the phenomenon to be modeled, not merely removable noise
\citep{pavlick-kwiatkowski-2019-inherent,plank-2022-problem}.  We retain reader-level trajectories and report
both group-level reliability and a leave-one-reader-out human--human
comparison, so a model score can be interpreted relative to actual reader
agreement.

LLMs can perform competitively as inexpensive annotators on some categorical
tasks \citep{gilardi-etal-2023-chatgpt}, and comparative prompting can improve
the reliability of automatic emotion-intensity labels
\citep{bagdon-etal-2024-expert}.  However, surveys and distributional
comparisons show that success on aggregate classification does not establish
faithful representation of divergent human opinions
\citep{pavlovic-poesio-2024-effectiveness}.  LLMs have also been proposed as
``silicon samples'' conditioned to emulate human subpopulations
\citep{argyle2023out}.  Later evaluations caution that
aggregate plausibility does not guarantee faithful simulation of diverse or
profile-specific responses \citep{yu-etal-2025-analysis}; even human approval
of model-suggested labels can shift subjective label distributions
\citep{schroeder-etal-2025-human-loop}.  Accordingly, our
task does not assume that a model \emph{is} a reader.  It tests the narrower,
measurable claim that its future-blind VAD trajectory aligns with the
collective trajectory and transition pattern reported by human readers.

%% file: sections/dataset.tex
\paragraph{Corpus and segmentation.}
LitArc-Bench comprises 39 Chinese-language excerpts and preserves the seven
consecutive segment boundaries used in the classroom reading activity. It is
designed as a high-resolution diagnostic benchmark: each excerpt is observed
at seven successive checkpoints and is supported by repeated judgments from
multiple readers. The corpus spans nine excerpts originally written in Chinese
and 30 Chinese translations. The translated subset comprises 23 works of
fiction and seven works of nonfiction, memoir, or criticism. Source
languages are Chinese (9), English (25), French (2), German (1), Japanese (1),
and Norwegian (1).

The benchmark does not resegment texts after observing the ratings. Segment
length has a median of 99 Unicode characters (interquartile range 68--145),
while complete excerpts have a median length of 747 characters (interquartile
range 697--794). The full ranges are 7--424 characters per segment and
529--1,395 per excerpt. Several checkpoints are therefore intentionally short:
they correspond to the units on which participants actually made judgments,
not to length-normalized model inputs.

\begin{table}[t]
\centering
\small
\begin{tabular}{lr}
\toprule
Statistic & Value \\
\midrule
Excerpts / segments per excerpt & 39 / 7 \\
Original Chinese / translated & 9 / 30 \\
Participants & 81 \\
Reader--text questionnaires & 430 \\
Raters per text, mean [range] & 11.03 [5, 17] \\
Segment-level rows & 3,010 \\
Observed / missing VAD cells & 9,020 / 10 \\
\bottomrule
\end{tabular}
\caption{Core corpus and annotation statistics.}
\label{tab:dataset-stats}
\end{table}

\paragraph{Participants and rating task.}
Data were collected from 81 adult participants across two classroom cohorts
at a Beijing university in June 2026. Before rating, participants were informed orally that their
responses would be used for research. Participation was voluntary;
participation decisions did not affect course grades or ordinary
course-performance assessment, and no monetary payment was provided. The form
stated that all information would be processed anonymously and requested only
the assigned text id and binary gender, with no direct identifier; 43
participants marked female and 38 marked male. Participants rated each segment
immediately after reading it on three 1--9 VAD scales: valence, arousal, and
dominance.

The instructions defined all three dimensions, provided endpoint anchors, and
emphasized first impressions rather than a correct answer. Readers were told to
proceed segment by segment, not to read the complete excerpt before rating, and
not to revise earlier answers; the estimated duration was five minutes.
Excerpts were distributed during class. Participants completed between one and
ten questionnaires (median six; mean 5.31), producing 430 reader--text
questionnaires. This repeated-measures collection yields a mean of 11.03
reader judgments per excerpt and supports more precise estimation of each
seven-point consensus arc than a one-judgment-per-text design. The
supplementary materials include a transcription of the complete instructions
from the blank instrument.

\paragraph{Records, missingness, and provenance.}
The dataset contains 430 reader--text questionnaires and 3,010 segment-level
rows. Of 9,030 possible VAD cells, ten are missing; these cells remain missing
and are never imputed. Each text has between 5 and 17 raters, with a mean of
11.03. Primary consensus scores are arithmetic means of all available
responses for the relevant text, segment, and dimension.
The public artifact includes the underlying numeric ratings after a second
de-identification step: reader codes are reassigned independently within each
text, so the released identifiers do not encode a cross-text participant link.
Gender, original subject codes, and completed forms are not released.

Raw bibliographic source notes are recorded for all 39 excerpts and preserved
separately from normalized metadata so that visible transcription errors are
not silently converted into guessed bibliographic facts. All 39 source works
are identified and annotated with original language, broad genre, and source
type. This separation of original records, normalized fields, and documented
release decisions follows established recommendations for data statements and
dataset documentation \citep{bender-friedman-2018-data,gebru-etal-2021-datasheets}.
One excerpt is an unpublished in-house translation rather than a published
edition. Its normalized record identifies the original work and translator
without attributing the text to a publisher, and its passage text is excluded
from the public release.

%% file: sections/human_reference.tex
We estimate inter-reader reliability separately for each text and dimension
using intraclass correlation and repeated split-half agreement. We report the
two-way random-effects, absolute-agreement reliability of the mean panel,
ICC(2,$k$), following standard ICC definitions and reporting guidance
\citep{shrout-fleiss-1979-icc,mcgraw-wong-1996-icc,koo-li-2016-icc}. On the
full, unfiltered ratings, the median ICC(2,$k$) is 0.531 for valence, 0.262 for
arousal, and 0.348 for dominance. This dimensional asymmetry motivates treating
VAD as partially distinct components of reader response rather than as three
equally reliable labels. Reliability is always estimated on the full panels
and is not improved by removing readers. We retain negative ICC estimates,
which occur for 4 valence, 11 arousal, and 5 dominance arcs. The corresponding
single-reader ICC(2,1) sensitivity analysis preserves the positive rank
associations between human reliability and model agreement ($\rho=.340$, .514,
and .370), indicating that the pattern is not an artifact of differing panel
sizes.

The primary human arc is the segment-wise mean of all available ratings for
each text and dimension. Our human--human reference uses a leave-one-reader-out
protocol. For each text and dimension, a reader's complete seven-point
trajectory is compared with the segment-wise mean trajectory of all other
complete readers. We summarize readers within each text before taking the
median across texts, preventing texts with larger panels from dominating the
benchmark. Confidence intervals use 10,000 bootstrap resamples of the 39 texts.
Negative correlations are retained, while correlations involving a constant
trajectory remain undefined. A screened, symmetric
drop-one-from-each-tail estimator is reported only as a sensitivity analysis;
it is not used as the primary target because screening by agreement with the
group could inflate apparent model alignment.

Split-half agreement uses 1,000 balanced random partitions per text and
dimension. When the number of complete readers is odd, one reader is randomly
left unused in each partition.

\begin{table*}[t]
\centering
\scriptsize
\setlength{\tabcolsep}{4pt}
\begin{tabular}{lrrrrr}
\toprule
Dimension &
\shortstack{Typical reader $r$\\{[95\% CI]}} &
\shortstack{MAE\\{[95\% CI]}} &
\shortstack{Transition $r$\\{[95\% CI]}} &
\shortstack{Raw split-half $r$\\{[95\% CI]}} &
\shortstack{Reader--text\\pairs} \\
\midrule
Valence &
.516 [.414, .555] &
.987 [.909, 1.048] &
.395 [.303, .549] &
.597 [.470, .660] &
427 \\
Arousal &
.274 [.187, .352] &
1.169 [1.071, 1.273] &
.252 [.213, .378] &
.409 [.299, .477] &
426 \\
Dominance &
.330 [.181, .499] &
1.286 [1.130, 1.440] &
.343 [.175, .409] &
.411 [.295, .589] &
427 \\
\bottomrule
\end{tabular}
\caption{Empirical human--human trajectory reference. Typical-reader metrics
compare one reader with the mean trajectory of the remaining readers and
weight texts equally. Raw split-half correlations compare two observed reader
groups and are not Spearman--Brown corrected.}
\label{tab:human-reference}
\end{table*}

Table~\ref{tab:human-reference} shows that a typical reader follows the
remaining-reader consensus most closely for valence ($r=.516$), followed by
dominance ($r=.330$) and arousal ($r=.274$). The same ordering appears in
absolute error, while group-level split-half estimates are higher because
averaging suppresses individual variation. We therefore treat these values as
empirical reference distributions rather than hard ceilings: a model can be
closer to the group mean than a typical individual without reproducing the
distribution of individual responses.

%% file: sections/experiments.tex
We evaluate LLMs under an incremental-reading protocol matched to the human
sequential instructions. We use \emph{future-blind} to mean
\emph{prompt-future-blind}: at step $s$, a model has access only to segments
$1,\ldots,s$ and reports current integer VAD ratings on the human 1--9 scales.
Later segments are not exposed when earlier ratings are elicited.

The primary target is the segment-wise mean of all available human ratings for
the corresponding text and dimension. Missing cells affect only that
segment--dimension mean and are not imputed. We do not remove readers from the
primary target. This choice keeps model evaluation aligned with the full-panel
leave-one-reader-out human reference and avoids mechanically increasing
agreement by screening readers according to their similarity to the group. A
screened, symmetric drop-one-from-each-tail estimate is retained only as a
sensitivity analysis.

We evaluate GPT-5.6-sol, Claude Opus 4.8, and DeepSeek-V4-Pro with one frozen
future-blind sequential protocol. Each model uses one fresh context per
retained text. The runner reveals one new segment only after a valid VAD
object has been submitted for the previous segment. Thus the model retains the
preceding segments and its own answers within a text, as a sequential reader
would, but is never shown a future segment or human annotation. Fresh contexts
contain only the fixed task prompt and text segments revealed so far. The runner accepts only
\texttt{\{"V": int, "A": int, "D": int\}} with values from 1 to 9 and writes
each accepted response immediately. Each analyzed condition contains 273 valid
VAD vectors. All three use the same semantic prompt, information
boundary, segment order, scale anchors, output schema, and validation rules.
Model identifiers, access dates, prompts, and validation procedures are
provided in the supplementary materials.
This expanding-prefix design defines a common information boundary for
trajectory recovery; it is not a test of whether retained history itself
causes better predictions. That mechanism would require a matched
current-segment-only condition.

We report within-text trajectory correlation as the primary shape-sensitive
measure, together with pooled correlation, absolute error, signed bias, and
adjacent-segment change agreement. Confidence intervals use 10,000 bootstrap
resamples of texts.
Because each trajectory contains seven positions, we also
report Spearman correlation and recompute Pearson correlation seven times per
text, each time omitting one position. Model--control comparisons use paired
text-level bootstrap differences and exact two-sided sign tests; segment rows
are never treated as independent samples. Benjamini--Hochberg correction
controls the false discovery rate across the nine model--dimension
comparisons within each diagnostic family. We use GPT-5.6-sol as the focal condition for the deeper
human-reference, reliability, and qualitative diagnostics, while the matched
three-model comparison tests whether the overall tracking pattern extends
beyond a single system.

\paragraph{Shallow baselines.}
We construct three corpus-level baselines without using the target text's human
ratings. A global-mean baseline predicts the mean VAD score estimated from the
other 38 texts. A position-mean baseline predicts the corresponding segment
mean, again excluding the target text. A negative control randomly permutes
the seven values of this position arc 1,000 times. The constant baseline has no
defined within-text correlation but provides an absolute-error reference.

We also implement a zero-training lexical baseline. The distributed
simplified-Chinese version of Chinese EmoBank contains 5,465 entries with
1--9 valence and arousal norms \citep{lee-etal-2022-chinese-emobank}.
Maximum-forward matching selects the longest entry at each position; a segment
receives the mean of its matches, or the fixed scale midpoint (5) when none
match. The lexicon covers 267/273 segments. For dominance, we audit AANC, a
2,061-entry four-character VAD lexicon \citep{liu-etal-2021-aanc}. Because it
covers only 16/273 segments, it is a supplementary check rather than a primary
comparator.

\paragraph{Joint and text-identity controls.}
An additive leave-one-text-out V/A predictor fits segment-position fixed
effects and Chinese EmoBank on the other 38 texts. For the text-identity test,
we subtract this cross-fitted shallow prediction from the held-out human V/A
arc (the leave-one-text-out position prior for dominance), while leaving model
trajectories untouched. Each model trajectory is then correlated with the
residual human arc for its correct text and all 38 wrong texts. This
human-outcome-only residualization keeps the candidate human arcs on a common
scale and makes text labels exchangeable under the null. We report paired
matched-minus-mismatched medians, retrieval, and 20,000 text-label
permutations. As an implementation check, we replace every model trajectory
with the same corpus-average arc; this text-invariant control must return
chance identity evidence.

\paragraph{Crossed uncertainty.}
We supplement the text bootstrap with 10,000 crossed participant--text
Bayesian-bootstrap replicates \citep{rubin-1981-bayesian-bootstrap,
owen-2007-pigeonhole}. Participant weights are shared across every rated text;
separate text weights propagate stimulus uncertainty. Human means, position
priors, joint position--lexicon predictions, and identity advantages are
recomputed per replicate.

%% file: sections/results.tex
\paragraph{Human reference and focal recovery.}
The 430 repeated reader--text questionnaires provide dense human reference
arcs: every excerpt has 5--17 readers (mean 11.03). Agreement is strongly
dimension-dependent. The leave-one-reader-out analysis in
Table~\ref{tab:human-reference} confirms the ICC ordering: valence trajectories
are the most reproducible, whereas arousal exhibits the greatest individual
variation. Model correlations are therefore interpreted against the observed
reader distribution rather than an assumed perfect target.

\begin{table*}[t]
\centering
\scriptsize
\setlength{\tabcolsep}{4pt}
\begin{tabular}{lrrrrr}
\toprule
Dimension &
\shortstack{Median text $r$\\{[95\% CI]}} &
Pooled $r$ &
\shortstack{Median\\MAE} &
\shortstack{Transition $r$\\{[95\% CI]}} &
Mean bias \\
\midrule
Valence & .725 [.460, .814] & .766 & 1.286 & .692 [.524, .814] & $-.713$ \\
Arousal & .607 [.462, .700] & .676 & 1.039 & .574 [.365, .718] & $-.443$ \\
Dominance & .618 [.376, .658] & .599 & 1.179 & .505 [.423, .656] & $-.694$ \\
\bottomrule
\end{tabular}
\caption{GPT-5.6-sol future-blind agreement with the full-panel human
mean. Bias is model minus human; intervals resample the 39 texts.}
\label{tab:gpt-main}
\end{table*}

Table~\ref{tab:gpt-main} shows substantial text-specific trajectory tracking.
The median within-text correlations are .725 for valence, .607 for arousal, and
.618 for dominance. Agreement on adjacent changes follows the same ordering.
The negative biases indicate consistently lower numeric ratings: less pleasant
for valence, less activated for arousal, and less dominant for dominance.
These dimension-specific shifts should not be collapsed into a single claim
about ``affective intensity.''

\begin{table}[t]
\centering
\small
\begin{tabular}{lrrr}
\toprule
Baseline & V & A & D \\
\midrule
Global mean & -- & -- & -- \\
Position mean & .167 & .470 & .268 \\
Chinese EmoBank & .259 & .252 & -- \\
Permuted position & .013 & $-.032$ & $-.010$ \\
\bottomrule
\end{tabular}
\caption{Median within-text trajectory correlation for shallow baselines.
Chinese EmoBank supplies V/A but not D. A constant prediction has undefined
correlation; intervals and coverage are in the supplement.}
\label{tab:simple-baselines}
\end{table}

\paragraph{How much do shallow explanations recover?}
Table~\ref{tab:simple-baselines} reveals a substantial corpus-level position
effect for arousal: an arc estimated only from the other 38 texts reaches
.470. Across the corpus, mean arousal rises from 5.03 at the first segment to
5.71 at the sixth before ending at 5.64. Chinese EmoBank reaches only .259 for
valence and .252 for arousal despite matching 97.8\% of segments. A model must
therefore exceed explicit lexical affect and, especially for arousal, the
generic expectation of narrative buildup before its alignment provides
evidence of text-specific tracking.

The joint position--lexicon predictor reaches shape $r=.219/.316$ and
transition $r=.330/.279$ for V/A. All six model transition advantages
(.172--.333) have text-bootstrap intervals above zero; exact sign-test BH
$q<.05$ in three cells and $q=.053$ in three. Five of six shape intervals
exclude zero (Table~\ref{tab:joint-identity}). The joint baseline is nevertheless
better calibrated: MAE is .706/.491 and lower than every model in all six
paired comparisons.

\paragraph{Text-specific change beyond shallow controls.}
Paired comparisons sharpen this result. GPT's median trajectory-correlation
advantages over the position prior are .577 [.332, .688], .173 [.050, .378],
and .298 [$-.018$, .485] for valence, arousal, and dominance. The dominance
shape advantage is therefore less certain, but adjacent-change advantages are
positive in all three dimensions: .858 [.610, 1.145], .267 [.111, .461], and
.595 [.231, .724]. Seven-point rank correlations remain .709, .558, and .468,
and median correlations after omitting each position in turn are .738, .600,
and .607. Thus the headline pattern is not driven by one checkpoint or by
Pearson's linearity assumption; dominance is most defensibly characterized as
recoverable local change with less certain global shape advantage.

A stricter countertrend diagnostic asks whether models recover a human change
precisely when a shallow control points in the opposite direction. Selection
uses only the human and control trajectories, never model predictions. Among
transitions with absolute human change at least 0.5, the three models recover
the human direction on 69--76\% of 58 valence transitions that oppose the
position prior, 61--68\% of 41 arousal transitions, and 53--67\% of 43
dominance transitions. When Chinese EmoBank points the other way, accuracy is
61--68\% on 41 valence transitions and 54--73\% on 48 arousal transitions.
The pattern persists across 0.25 and 0.75 thresholds
(Appendix~\ref{tab:countertrend}). These cases directly isolate changes that
cannot be explained by following the corpus's average narrative position or
the local lexicon trajectory.

After cross-fitted removal of shallow effects from the human target, the
correct model--human pairing exceeds the median of 38 wrong texts by
.447--.573 in all nine cells; every interval is above zero and every identity
permutation has $p<.0001$ (Table~\ref{tab:joint-identity}). Top-1 retrieval is
7.7--20.5\% versus 2.6\% chance. Replacing every model trajectory with the
same corpus-average arc yields $p=1$ in every cell, confirming that the test
does not manufacture identity evidence from a generic trajectory.

\begin{figure*}[t]
\centering
\includegraphics[width=.96\textwidth]{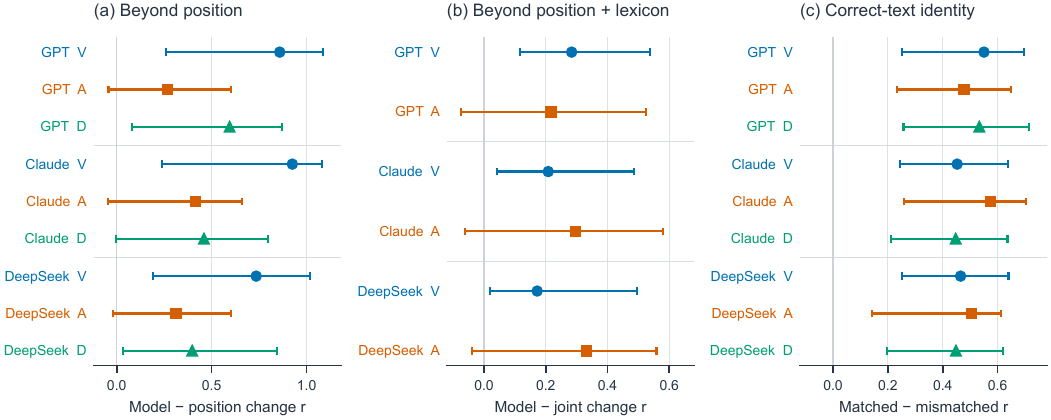}
\caption{Progressive controls on model--human trajectory recovery. Points are
medians and bars are crossed participant--text 95\% intervals. (a) Advantage
in adjacent-change correlation over the leave-one-text-out position prior.
(b) Advantage over the joint position--lexicon control; this panel contains
V/A only because Chinese EmoBank provides no dominance scores. (c)
Cross-fitted correct-text identity advantage, measured as the matched
correlation minus the median correlation with the other 38 human texts; the
human target is residualized against the joint control for V/A and the
position prior for D.}
\label{fig:diagnostics}
\end{figure*}

\paragraph{Human variability and calibration.}
GPT-5.6-sol exceeds the typical-reader reference on 69.2\%, 82.1\%, and
79.5\% of texts. The median paired differences
between model and typical-reader correlations are .166 [.031, .249], .321
[.185, .405], and .153 [.063, .236]. These positive differences mean that the
model often approximates the aggregate arc more closely than a single reader
does; they do not establish superhuman reading, because the model is evaluated
against the group mean whereas each human observation contains genuine
individual response variation.

Against Chinese EmoBank on the same texts, GPT's paired median advantages in
within-text correlation are .407 [.273, .515] for valence and .400 [.203,
.670] for arousal; its transition-correlation advantages are .259 [.166, .416]
and .236 [.139, .523]. The model thus tracks more than local affective lexical
density. It is not uniformly better calibrated: GPT-minus-lexicon MAE
differences are +.561 [.246, .867] and +.130 [$-.079$, .388]. A
midpoint-leaning lexical average can be numerically closer while recovering
less trajectory shape. The coverage-limited AANC dominance audit remains in
the supplement.

Figure~\ref{fig:diagnostics} and
Table~\ref{tab:full-cross-model-results} summarize the controlled three-model
comparison. Claude-minus-DeepSeek median shape differences are $-.007$, .018,
and .025 across VAD, with every interval crossing zero. Yet Claude has lower
MAE ($-.377/-.319/-.286$) and higher concordance
(.053/.107/.032); concordance penalizes departures from the identity line
\citep{lin1989ccc}. Thus all three matched conditions track trajectories, most
clearly for V/A shape and VAD changes, while their ordering depends on whether
the target is shape or calibration. Full intervals are in the supplement.

Against the position control, all nine median shape advantages are positive.
Valence replicates across models (.540--.607, intervals above zero); arousal
is supported for GPT and Claude but uncertain for DeepSeek, and dominance
shape intervals include zero. In contrast, all nine adjacent-change
advantages have intervals above zero (.267--.925) and FDR-corrected sign tests
remain significant (maximum $q=.024$; Table~\ref{tab:cross-model-position}).
Text-specific evidence is therefore strongest for valence shape and local VAD
change, not uniformly for global shape.

Under crossed participant--text reweighting, all nine cross-fitted identity
advantages retain intervals above zero. Joint-control shape and transition
intervals also exclude zero for valence across models. For arousal, posterior
mass above zero is .856--.969 for shape and .923--.954 for transitions, but
95\% intervals include zero (Table~\ref{tab:crossed-bootstrap}). Identity is
therefore robust to repeated-reader dependence; the joint-control arousal
comparison is directional under crossed uncertainty.

%% file: sections/analysis.tex
\subsection{Reliability and trajectory amplitude}
\label{sec:reliability-amplitude}

Raw ICC--alignment Spearman associations are .345/.520/.398 across VAD, but
become .011/.324/.114 after controlling ranked human-arc SD. Only arousal has
residual-permutation $p<.05$ ($p=.045$), and its bootstrap interval includes
zero, indicating that amplitude explains much of the raw valence and
dominance association. Positive
model--typical-reader gaps (.166/.321/.153) reflect aggregation, not
superhuman reading: the model targets a panel mean, whereas individual
agreement includes response variation. Full intervals and an 18-case
qualitative shortlist accompany the artifact.

%% file: sections/conclusion.tex
LitArc-Bench separates position and lexical affect from text-specific
reader-response dynamics. Across 39 excerpts, three future-blind models recover
correct-text structure beyond shallow controls: crossed resampling preserves
all nine identity and all valence joint-control advantages, with weaker
arousal evidence. Recovery differs by dimension and can coexist with scale
bias, motivating separate evaluation of recovery, calibration, and human
variation.

%% file: sections/limitations.tex
LitArc-Bench is a high-resolution diagnostic collection, not a population
survey of Chinese literature or readers. Its 39 excerpts retain the original
seven reading checkpoints, including naturally varying segment lengths, and
its 81 adults were recruited from two cohorts at one university. Most
participants rated multiple excerpts (median six), yielding dense
reader--text coverage but 81 independent people and one recruitment site.
Excerpts were distributed in class, but the preserved record does not support
claims of randomized allocation, counterbalancing, or order/fatigue analysis.
Primary analyses aggregate within excerpts and weight excerpts equally;
crossed participant--text resampling additionally shares each participant's
weight across every rated text and recomputes the human targets and position
priors, joint controls, and identity comparisons. Claims concern this
controlled benchmark rather than population
prevalence.

LLM outputs may vary with model version, prompting, decoding configuration, and
provider-side changes. We record model identifiers, access dates, prompts,
available run configurations, and outputs, but each condition contains one run
per text and should be interpreted as a fixed snapshot; reported uncertainty
does not include run-to-run or provider-side variation. Genre- and source-type
analyses are exploratory. The high-coverage Chinese lexical comparator supplies
only valence and arousal; AANC dominance norms match 6.8\% of segments and
therefore serve only as a coverage audit. In addition, the human dominance
instruction explicitly invites situational perspective-taking, whereas the
model prompt uses the same powerless/control anchors without that clause;
dominance comparisons are consequently interpreted most cautiously.

The shallow controls establish information beyond corpus-level position trends
and lexical-affect averages. Our claim is therefore about text-specific
trajectory recovery under a future-blind boundary, not the causal benefit of
context accumulation. We do not include a same-model current-segment-only
ablation and cannot isolate the contribution of retained prior context. Future
passages are absent from the prompt, but models may still recognize well-known
works from pretraining; the protocol prevents prompt-level look-ahead, not
parametric knowledge of a story.

Public reproducibility is intentionally tiered: aggregate and text-local
reader analyses are recomputable; crossed resampling is auditable through
code, schema, seed, self-test, and frozen replicate summaries but requires
private cross-text linkage to rerun; prompt reconstruction requires lawful
access to the cited editions.

%% file: sections/ethics.tex
Before rating, adult students were informed orally that their responses would be used for
research and voluntarily agreed to participate. Participation decisions did
not affect course grades or ordinary course-performance assessment, and no
monetary payment was provided. The blank questionnaire explicitly stated that all
information would be processed anonymously, collected no direct identifiers,
and described the task as having no standard answer. The institution confirmed
through its applicable procedure that formal ethics review was not required
for this anonymous, minimal-risk classroom study; no approval or exemption
identifier was issued.

The public release follows data minimization. It includes the numeric rating
rows after reader codes are reassigned independently within each text, together
with aggregate segment statistics, consensus trajectories, bibliographic
metadata, normalized model outputs, baseline intermediates, and evaluation
code. This preserves reproducibility of the human reference without exposing
an identifier-based cross-text participant link. Stable study codes in the
internal digitized record support participant-level crossed resampling; their
assignment procedure is not present in the recovered blank instrument or
preserved procedure record, and the codes and cross-text mapping remain
private. To the extent that the
project holds the relevant rights, de-identified numeric ratings, aggregate
annotations, metadata, documentation, and result tables are CC BY 4.0, while
code is MIT-licensed. Original subject codes, gender responses, and completed
paper forms are not publicly distributed. The institution confirmed
that these de-identified reader-level numeric ratings may be publicly shared
under this data-minimization strategy; the responsible faculty member and all
authors also reviewed and confirmed the release. The project grants no
license for third-party literary text.
Literary passages are omitted; the release instead provides provenance,
segment hashes, and a local alignment mechanism for researchers with lawful
access to the cited source. The unpublished in-house translation is designated
derived-annotations-only and is likewise absent from the passage-text package.

LitArc-Bench is intended for research on aggregate literary-response
trajectories, not for inferring an individual's mental state, grading students,
or replacing human readers in educational or psychological assessment. Model
agreement with a small classroom panel may encode culturally and
demographically narrow response patterns; consequently, the reported scores
should not be interpreted as universal norms of literary experience.

%% file: sections/appendix.tex
\section{Rating Instrument}
\label{app:instrument}

The original blank Chinese questionnaire is preserved with the study records.
It states that responses are anonymous and asks only for the assigned excerpt
identifier and binary gender. The following is an English transcription of
the task instructions:

\begin{quote}
Read a Chinese-language literary excerpt in seven consecutive segments.
Immediately after each segment, rate the feeling evoked in you on valence (V),
arousal (A), and dominance (D). Use your first impression; there is no correct
answer. Do not read ahead or revise an earlier answer.
\end{quote}

Each dimension uses an integer scale from 1 to 9. The displayed endpoint and
midpoint anchors are:

\begin{center}
\small
\resizebox{\columnwidth}{!}{%
\begin{tabular}{llll}
\toprule
Dimension & 1 & 5 & 9 \\
\midrule
Valence & extremely unpleasant & neutral & extremely pleasant \\
Arousal & extremely calm & moderate & extremely activated \\
Dominance & extremely powerless & moderate & extremely in control \\
\bottomrule
\end{tabular}
}
\end{center}

Arousal is explicitly defined as independent of valence. Dominance asks
whether the reader feels powerless or suppressed versus powerful or in
control. The same three response rows follow every segment.

\section{Model Elicitation and Validation}
\label{app:prompt}

\paragraph{Primary future-blind condition.}
GPT-5.6-sol was accessed on July 27, 2026 through a Codex subagent interface
with low reasoning effort; temperature was not exposed. One fresh context was
used for each of the 39 retained texts. At each step, the session contained the
cumulative textual prefix and the human
form's three 1--9 scale definitions, and the runner required one object of the form
\texttt{\{"V": integer, "A": integer, "D": integer\}}.
The next segment appeared only after validation; no human rating or information
about the analysis was supplied.

\paragraph{Matched model conditions.}
Claude Opus 4.8 and DeepSeek-V4-Pro used the same frozen future-blind runner as
GPT-5.6-sol. Each received one new segment only after its preceding VAD object
was validated and saved. The artifact normalizes all three outputs to one
schema and omits provider transcripts and hidden reasoning. The retained
prefix is the common evaluation boundary; no claim is made here that it
outperforms a current-segment-only condition.

\section{Additional Quantitative Results}
\label{app:full-results}

\begin{table}[h]
\centering
\resizebox{\columnwidth}{!}{%
\begin{tabular}{lrrrr}
\toprule
Dim. & Spearman & Leave-one-position $r$ &
\shortstack{Paired $r$ advantage\\{[95\% CI]}} &
\shortstack{Paired transition advantage\\{[95\% CI]}} \\
\midrule
V & .709 & .738 & .577 [.332, .688] & .858 [.610, 1.145] \\
A & .558 & .600 & .173 [.050, .378] & .267 [.111, .461] \\
D & .468 & .607 & .298 [$-.018$, .485] & .595 [.231, .724] \\
\bottomrule
\end{tabular}
}
\caption{Robustness of GPT-5.6-sol relative to the leave-one-text-out position
prior. Leave-one-position values summarize seven six-point correlations per
text. Intervals resample texts and preserve pairing.}
\label{tab:trajectory-robustness}
\end{table}

\begin{table}[h]
\centering
\resizebox{\columnwidth}{!}{%
\begin{tabular}{llrrrr}
\toprule
Model & Dim. &
\shortstack{Median $r$\\{[95\% CI]}} &
\shortstack{Median CCC\\{[95\% CI]}} &
\shortstack{Median MAE\\{[95\% CI]}} &
\shortstack{Transition $r$\\{[95\% CI]}} \\
\midrule
Claude & V & .636 [.547, .764] & .318 [.151, .444] & 1.024 [.798, 1.163] & .661 [.533, .744] \\
       & A & .628 [.558, .791] & .339 [.189, .472] & .844 [.733, .939] & .587 [.398, .651] \\
       & D & .490 [.332, .633] & .158 [.106, .300] & .964 [.765, 1.317] & .460 [.239, .579] \\
\addlinespace
DeepSeek & V & .629 [.449, .730] & .210 [.136, .340] & 1.357 [1.012, 1.494] & .611 [.427, .668] \\
         & A & .573 [.482, .689] & .242 [.188, .276] & 1.163 [.987, 1.345] & .617 [.420, .658] \\
         & D & .416 [.310, .612] & .164 [.076, .252] & 1.369 [1.000, 1.673] & .524 [.283, .628] \\
\bottomrule
\end{tabular}
}
\caption{Additional future-blind model results with text-bootstrap 95\%
intervals; all conditions share the frozen sequential boundary.}
\label{tab:full-cross-model-results}
\end{table}

\begin{table}[t]
\centering
\scriptsize
\setlength{\tabcolsep}{3.5pt}
\resizebox{\columnwidth}{!}{%
\begin{tabular}{llrrr}
\toprule
Model & Dim. &
\shortstack{Shape advantage\\{[95\% CI]}} &
\shortstack{Positive\\texts} &
\shortstack{Transition advantage\\{[95\% CI]}} \\
\midrule
\multirow{3}{*}{GPT-5.6-sol}
 & V & .577 [.332, .688] & 79.5\% & .858 [.610, 1.145] \\
 & A & .173 [.050, .378] & 71.8\% & .267 [.111, .461] \\
 & D & .298 [$-.018$, .485] & 64.1\% & .595 [.231, .724] \\
\addlinespace
\multirow{3}{*}{Claude}
 & V & .607 [.316, .749] & 76.3\% & .925 [.485, 1.222] \\
 & A & .292 [.087, .431] & 74.4\% & .414 [.058, .529] \\
 & D & .148 [$-.006$, .412] & 60.5\% & .459 [.334, .590] \\
\addlinespace
\multirow{3}{*}{DeepSeek}
 & V & .540 [.306, .671] & 79.5\% & .734 [.502, 1.167] \\
 & A & .100 [$-.009$, .378] & 61.5\% & .312 [.168, .447] \\
 & D & .172 [$-.004$, .369] & 64.1\% & .398 [.316, .571] \\
\bottomrule
\end{tabular}
}
\caption{Paired advantages over the leave-one-text-out position prior.
Intervals use 10,000 text bootstrap resamples. Shape is the difference in
seven-point trajectory correlation; transition uses the six adjacent changes.
All nine transition sign tests survive Benjamini--Hochberg correction
($q<.05$); exact $p$ and $q$ values are in the released result table.}
\label{tab:cross-model-position}
\end{table}

\begin{table}[t]
\centering
\scriptsize
\setlength{\tabcolsep}{3pt}
\resizebox{\columnwidth}{!}{%
\begin{tabular}{llrrr}
\toprule
Model & Dim. &
\shortstack{Shape advantage\\over joint [95\% CI]} &
\shortstack{Transition advantage\\over joint [95\% CI]} &
\shortstack{Matched residual\\advantage [95\% CI]} \\
\midrule
\multirow{2}{*}{GPT-5.6-sol}
 & V & .488 [.293, .645] & .284 [.174, .408] & .550 [.427, .643] \\
 & A & .153 [.023, .366] & .217 [.047, .439] & .477 [.408, .641] \\
\addlinespace
\multirow{2}{*}{Claude}
 & V & .406 [.263, .569] & .208 [.072, .349] & .452 [.358, .570] \\
 & A & .202 [.040, .473] & .297 [.108, .461] & .573 [.349, .682] \\
\addlinespace
\multirow{2}{*}{DeepSeek}
 & V & .373 [.230, .609] & .172 [.039, .404] & .465 [.371, .623] \\
 & A & .159 [$-.031$, .262] & .333 [.078, .448] & .505 [.281, .644] \\
\bottomrule
\end{tabular}
}
\caption{Joint shallow-control and matched-text results. The additive
leave-one-text-out baseline combines position and Chinese EmoBank. Matched
residual advantage subtracts the median correlation with the 38 wrong
cross-fitted human residual arcs; model arcs are not residualized. Dominance
matched advantages, using position control, are .533, .447, and .448 for GPT,
Claude, and DeepSeek. All nine identity permutation tests have $p<.0001$;
text-invariant negative controls have $p=1$.}
\label{tab:joint-identity}
\end{table}

\begin{table}[t]
\centering
\scriptsize
\setlength{\tabcolsep}{4pt}
\begin{tabular}{llrr}
\toprule
Model & Dim. &
\shortstack{Crossed identity\\advantage [95\% CI]} &
\shortstack{Joint transition\\advantage [95\% CI]} \\
\midrule
\multirow{3}{*}{GPT-5.6-sol}
 & V & .550 [.250, .697] & .284 [.116, .537] \\
 & A & .477 [.233, .649] & .217 [$-.074$, .524] \\
 & D & .533 [.257, .713] & -- \\
\addlinespace
\multirow{3}{*}{Claude}
 & V & .452 [.243, .636] & .208 [.041, .485] \\
 & A & .573 [.258, .702] & .297 [$-.062$, .580] \\
 & D & .447 [.212, .636] & -- \\
\addlinespace
\multirow{3}{*}{DeepSeek}
 & V & .465 [.250, .640] & .172 [.018, .495] \\
 & A & .505 [.141, .611] & .333 [$-.039$, .559] \\
 & D & .448 [.197, .619] & -- \\
\bottomrule
\end{tabular}
\caption{Crossed participant--text Bayesian-bootstrap robustness. Participant
weights are shared across all questionnaires; human targets, cross-fitted
controls, and comparisons are recomputed in 10,000 replicates. Identity
intervals exclude zero in all nine cells. Joint-control V intervals exclude
zero; A intervals preserve positive point estimates but include zero.}
\label{tab:crossed-bootstrap}
\end{table}

\begin{table}[t]
\centering
\scriptsize
\setlength{\tabcolsep}{4pt}
\begin{tabular}{llrrrr}
\toprule
Control & Dim. & $N$ & GPT & Claude & DeepSeek \\
\midrule
\multirow{3}{*}{Position}
 & V & 58 & .759 & .690 & .707 \\
 & A & 41 & .610 & .683 & .659 \\
 & D & 43 & .674 & .535 & .651 \\
\addlinespace
\multirow{2}{*}{Chinese EmoBank}
 & V & 41 & .683 & .610 & .634 \\
 & A & 48 & .542 & .646 & .729 \\
\bottomrule
\end{tabular}
\caption{Countertrend direction accuracy. Transitions are selected
independently of model predictions when the human change has magnitude at
least 0.5 and opposes the shallow control; ties count as incorrect. The
artifact includes text-block bootstrap intervals and sensitivity analyses at
0.25 and 0.75.}
\label{tab:countertrend}
\end{table}

\begin{table}[t]
\centering
\small
\begin{tabular}{lrrr}
\toprule
Dim. & Raw $\rho$ &
\shortstack{Partial $\rho$\\controlling arc SD} &
$p_{\mathrm{perm}}$ \\
\midrule
V & .345 & .011 [$-.291$, .322] & .948 \\
A & .520 & .324 [$-.038$, .612] & .045 \\
D & .398 & .114 [$-.195$, .428] & .491 \\
\bottomrule
\end{tabular}
\caption{Association between ICC(2,$k$) and focal-model trajectory agreement
before and after controlling human-arc amplitude. Intervals resample texts;
two-sided $p$ values use 50,000 residual permutations.}
\label{tab:reliability-amplitude}
\end{table}

\paragraph{Lexical coverage audit.}
Chinese EmoBank matches 267/273 segments and supplies valence and arousal.
AANC matches 16 segments; after neutral midpoint fallback, correlation is
defined for only 12 texts (11 for dominance with all-participant norms).
Dominance is therefore not used as a primary lexical baseline.

\section{Public Artifact Contents}
\label{app:artifact}

The artifact contains de-identified reader-level numeric ratings, text-free
aggregates, metadata, hashes, baseline intermediates, normalized predictions,
robustness results, a protocol audit, and scripts that verify the package,
locate exact segments in lawful local source text, execute the provider-neutral
future-blind state machine, and recompute human and model metrics. Code is
MIT-licensed. To the extent that the project holds the relevant rights,
de-identified numeric ratings, aggregates, metadata, documentation, and result
tables are CC BY 4.0. The artifact excludes
original subject codes, cross-text participant links, forms, gender,
unlicensed passages, credentials, provider transcripts, and hidden reasoning.